\documentclass[final,5p,times,twocolumn]{elsarticle}
\usepackage{amssymb,amsmath,amsthm,graphicx,latexsym,color}
\usepackage{tikz}
\usepackage{pgfplots}
\usepackage{doi}

\theoremstyle{definition}
\newtheorem{theorem}{Theorem}[]
\newtheorem{example}[theorem]{Example}

\newtheorem{definition}[theorem]{Definition}

\newtheorem{heuristic}{Heuristic}

\newcommand{\valid}{\mathsf{valid}}
\newcommand{\Valid}{\mathsf{Valid}}
\newcommand{\Traces}{\mathsf{Traces}}

\newcommand{\dparticipation}{\mathsf{Participation}}
\newcommand{\dinitial}{\mathsf{Initial}}
\newcommand{\dfinal}{\mathsf{End}}
\newcommand{\dsucc}{\mathsf{Succ}}
\newcommand{\dnotsucc}{\mathsf{NotSucc}}
\newcommand{\dresp}{\mathsf{Resp}}
\newcommand{\dchainresp}{\mathsf{ChainResp}}
\newcommand{\dweakresp}{\mathsf{WeakResp}}
\newcommand{\dprec}{\mathsf{Prec}}

\newcommand{\drespexist}{\mathsf{RespondExist}}

\newcommand{\Const}{\mathsf{Const}}
\newcommand{\MinConst}{\mathsf{MinConst}}

\newcommand{\Perms}{\mathsf{Perms}}

\newcommand{\effect}{\mathbf{comb\_diversity}}

\renewcommand{\P}{\mathbb{P}}
\newcommand{\Prob}{\P}
\definecolor{mygreen}{RGB}{80,160,80}
\definecolor{myonegrey}{RGB}{30,30,30}
\definecolor{mytwogrey}{RGB}{70,70,70}
\definecolor{mythreegrey}{RGB}{110,110,110}
\definecolor{myfourgrey}{RGB}{150,150,150}
\newcommand\mps[1]{}

\newcommand{\GConst}{\mathsf{GoodConst}}
\newcommand{\GValid}{\mathsf{GoodValid}}
\newcommand{\gvalid}{\mathsf{gvalid}}
\newcommand{\goodness}{\mathsf{goodness}}
\newcommand{\loggoodness}{\mathsf{log\_goodness}}
\newcommand{\pc}{\mathsf{pattern\_div}}
\newcommand{\npc}{\mathsf{norm\_pattern\_div}}

\newcommand{\dG}{\mathsf{G}}
\newcommand{\dF}{\mathsf{F}}
\newcommand{\dmodels}{\models}
\newcommand{\dimplies}{\Rightarrow}
\newcommand{\firstpass}{\mathsf{fp}}

\journal{}

\begin{document}

\begin{frontmatter}

\title{Combinatorial diversity metrics for the analysis of policy processes}

\date{}

\author[label2]{Mark Dukes}
\ead{mark.dukes@ucd.ie}
\author[label1]{Anthony A. Casey}
\ead{tony.casey@ucdconnect.ie}

\address[label2]{School of Mathematics and Statistics, University College Dublin, Ireland.}
\address[label1]{UCD School of Politics and International Relations, University College Dublin, Ireland.}

\begin{abstract}
We present several completely general diversity metrics to quantify the problem-solving capacity of any public policy decision making process. 
This is performed by modelling the policy process using a declarative process paradigm in conjunction with constraints modelled by expressions in linear temporal logic.
We introduce a class of traces, called first-passage traces, to represent the different executions of the declarative processes.
Heuristics of what properties a diversity measure of such processes ought to satisfy are 
used to derive two different metrics for these processes in terms of the set of first-passage traces.
These metrics turn out to have formulations in terms of the entropies of two different random variables on the set of traces of the processes.
In addition, we introduce a measure of `goodness' whereby a trace is termed {\it{good}} if it satisfies some prescribed linear temporal logic expression. 
This allows for comparisons of policy processes with respect to the prescribed notion of `goodness'.
\end{abstract}


\end{frontmatter}

\section{Introduction}
We present several completely general diversity metrics to quantify the problem-solving capacity of any public policy decision making process. 
We do this by modelling the public policy process using the declarative process paradigm originally developed in the the fields of information science and business process management (BPM).
Our approach differs markedly from current approaches to the modelling of the public policy process (cf.~\cite{WEIBLE}).

Although diagrammatic notions of the public policy process as a `policy cycle' emerged as long ago as the 1950’s~\cite{LASSWELL}, 
this conceptualisation never converged with the BPM graphical notations that evolved elsewhere in the business process re-engineering literature~\cite{DAVENPORT}. 
One consequence is that public policy process research has become detached from the results of the large body of process analysis research in the information science and business administration literatures.
A notable exception is the `garbage can' theory of organizational choice where there have been periodic attempts to model it more formally as a Petri net~\cite{HEITSCH}.
Another consequence is that much of the current business process research does not address the concerns of public administration and public policy process research. 

For example, a large literature has arisen around the process mining of computer log files generated by highly automated business processes (cf.~\cite{vda2014}). 
In contrast, the public policy decision making process, although certainly computer assisted, is a highly complex, largely manual process that generates few, or no, computer log files that it would be possible to mine. 

Similarly, there is now an active, predominantly imperative, business process metrics literature that originated in software maintenance and social network analysis (SNA) metrics research.
This strand of research has found applications in the analysis of public service delivery processes but not in the analysis of the generally much more complex public policy making processes. 
Comprehensive reviews of the literature surrounding the predominantly imperative business process metric research can be found in Gonz\'alez et al.~\cite{GONZALEZ}, Mendling~\cite[pp. 114--117]{MENDLING}  and Melcher~\cite[pp. 27--56]{MELCHER}.

In this paper we will take as our starting point the paradigm of a declarative process. 
This declarative approach involves declaring relations between activities in the policy process that may (or may not) happen,
and then studying the possible ways (termed {\it{traces}}) in which a policy process may be executed.
A natural setting for studying such declarative models is linear temporal logic (LTL),
an extension to propositional logic that includes temporal operators.
A useful tool in BPM is the Declare language~\cite{declarelanguage} which was developed for modelling LTL expressions.
The semantics of Declare is clearer than the corresponding LTL expressions and we adopt the Declare expressions and operators.
A common feature of the use of LTL is the appearance of infinite traces.
Research in the area has looked at a finite counterparts to LTL that deal with finite traces where this has been needed at an application level for tasks such as specification and verification~\cite{ltlfinite,finiteltlone}.

In order to overcome the issue of infinite traces, in this paper we will look at {\it{first-passage traces}} (defined in Section 2).
These traces capture two aspects of the processes which we deem to be important for our considerations: the order in which activities first happen and whether an activity eventually happens.

If, instead, we were to truncate the traces at some finite length and analyse those initial segments
then we will be losing information regarding how an activity that has not yet been seen is related to those that have appeared.
Indeed it may not have appeared at all in the possibly infinite trace, or it may have been waiting for another event to trigger it.

In choosing first-passage traces to be our representatives we could potentially be discarding information related to the medium-term temporal dynamics of the policy process.
However, on balance with other considerations for potential trace representatives, these first-passage traces are the most important for our current purposes.

Our modelling code (written in the SageMath computer algebra system) used a combination of reduction techniques in being able to compute the valid traces of a given declarative process. A discussion of these would be somewhat out of place in the current paper, but one overarching fact is that there is a combinatorial explosion with every extra activity introduced into a declarative process. The main reason for this is that once there is one more degree of freedom in when activities may occur for the first time with respect to one another, this will allow for a significant increase in the number of traces that will satisfy the constraint(s).

This paper is a first study of diversity measures in policy process analysis via the declarative paradigm.
As such, while some of our measures might appear crude at first glance, their derivation and introduction are strongly motivated as solutions to the heuristics we deem important to their existence.

In Section 2 we will introduce declarative processes and concepts to be used.
In Section 3 we use the first-passage traces of declarative processes along with several heuristics 
to derive a metric for comparing declarative processes and, in turn, the policies they model.
In Section 4 we introduce two metrics to measure the `goodness' of a declarative process with  respect to some LTL formula that serves as an indicator function for `goodness'. 
In Section 5 we discuss entropy in relation to the combinatorial diversity metric of Section 3 and see how the more general combinatorial diversity metric is in fact the entropy of a simple random variable on the space of first-passage traces.
In Section 6 we introduce a metric that is motivated by the distribution of permutation patterns in the traces of a declarative process.
This is another entropy measure and is invariant under the labelling of the set of activities.
It provides a measure of how free the collection of traces of a declarative process is in terms a specified resolution parameter.

\section{Declarative processes}
First let us introduce some standard notions related to process theory~\cite{vda}.
Let $\Sigma$ be a set of {\it{activities}} and let $\Sigma^{*}$ be the set of all sequences over $\Sigma$.
A {\it{trace}} is a sequence of activities $\sigma=(e_1,\ldots,e_n)\in \Sigma^{*}$ and we use $\epsilon$ to denote the empty trace.
An {\it{event}} is an occurrence of an activity in a trace.
A {\it{log}} is a multiset consisting of traces.\mps{log needed?}

A declarative constraint is a constraint on activities in a process.
By way of an example, given two activities $a$ and $b$ in $\Sigma$, we may wish to specify that event $b$ must happen as a response to event $a$.
In LTL one would represent our example preference by the LTL formula $\dG (a \dimplies \dF b)$, which can be read as
`it is globally true that ($a$ occurs implies $b$ occurs at some point after $a$)''.
The syntax for Declare is easier to deal with in this respect and uses $\dresp(a,b)$ for $\dG (a \dimplies \dF b)$.
A list of some popular Declare expressions is given in Figure~\ref{deccondefs}.

We say that a trace $\sigma$ satisfies the constraint $\dresp(a,b)$ if any occurrence of $a$ in the trace will feature an occurrence of $b$ to its right. To represent this we write $\sigma \dmodels \dresp(a,b)$.
It may be the case that $a$ and $b$ are not events in $\sigma$, in which case $\sigma$ certainly satisfies the constraint $\dresp(a,b)$.

As a further example consider the trace $\sigma = (3,3,2,4,1,4)$ with $\Sigma=\{1,2,3,4,5\}$.
The trace $\sigma$ satisfies the declarative constraint $\dresp(2,1)$, i.e. $\sigma \dmodels \dresp(2,1)$ since event $1$ happens after event $2$ in $\sigma$.
However, both $\sigma \dmodels \dresp(2,3)$ and $\sigma \dmodels \dresp(2,5)$ are false.

\begin{definition}
A {\it{declarative process}} is a process on a set of activities $\Sigma$ that satisfies all conditions in a set $\Const$ of declarative constraints.
We will represent this as a pair $D=(\Sigma,\Const)$.
The set of traces of the process is
$$\Traces(D) ~=~ \{ \sigma\in \Sigma^{*} ~:~ \sigma \dmodels c \mbox{ for all } c \in \Const\}.$$
\end{definition}
Restrictions on the beginning and ending of these processes may be incorporated into the constraint set using declarative constraints.

\begin{figure}\label{deccondefs}
\small
\centerline{
\begin{tabular}{|l|l|} \hline
    Constraint          & Explanation  \\ \hline \hline
    $\dparticipation(a)$ & $a$ occurs at least once \\ \hline
    $\dinitial(a)$      & event $a$ is first to occur   \\ \hline
    $\dfinal(a)$            & event $a$ is last to occur  \\ \hline
    $\dresp(a,b)$       & If $a$ occurs, then $b$ occurs after $a$ \\ \hline
    $\dchainresp(a,b)$  & If $a$ occurs, then $b$ occurs \\ &  immediately after $a$\\ \hline
    $\dprec(a,b)$       & $b$ occurs only if preceded by $a$ \\ \hline
    $\dsucc(a,b)$       & $a$ occurs iff it is followed by $b$ \\ \hline
    $\dnotsucc(a,b)$    & $a$ can never occur before $b$ \\ \hline
    $\dweakresp(a,b)$   & If $a$ occurs, then $b$ might occur \\ & after it \\ \hline
\end{tabular}}
\caption{Some typical Declare constraints}
\end{figure}

As mentioned in the introduction, the traces that we will consider are different.
For general declarative processes, traces of infinite length may occur.
Infinite traces are inconvenient when it comes to analysing the systems that a declarative processes is modelling, particularly if that system is known to be finite to be begin with.

We have given some reasons in the introduction for choosing a new type of trace (called a {\it{first-passage trace}}) that is different in spirit to those seen in finite versions of LTL.
In essence, the nature of what we are modelling (policies) is such that once an event occurs in a trace then we may continue to think of what it represents as being active throughout the remainder of the process.
Combining this with a desire to study the variety of ways in which events may occur in a declarative process, we settled upon
first-passage traces as reasonable representatives of the systems we are analyzing.
This idea of first-passage events is not new and has its motivation in models in applied probability where one is interested in the first time that a particular event occurs, the so called {\it{first-passage phenomena}}~\cite{render}.

The assumption that first-passage traces are good representatives is, of course, open to criticism.
An argument could be made for considering traces of a more general type.
However, in this first paper on the topic we will restrict our attention to these first-passage traces.
An advantage of this this assumption is that the length of traces is bounded by the size of the activity set.
This has allowed us to perform an analysis of systems consisting of up to 15 activities that have many relations between them.\mps{CITE COVID}
The number of traces one finds in these systems is typically very large and their derivation requires significant computational effort.

To add some perspective to this: the number of first-passage traces of a constraint-free declarative process consisting of 10 activities will be
9,864,102 traces. If we were to consider the traces of this system and make them finite by truncating the first 10 entries, then there will be $(10^{11}-1)/9 \sim 11,111,111,111$ traces.

In a first-passage trace we only record the first occurrence of an event.

\begin{definition}
Given a (possibly infinite) sequence $x=(x_1,x_2,\ldots) \in \Sigma^{*}$, let
$\firstpass(x)$ be the sequence that records the order in which the elements of $\Sigma$ first appear in $x$.
\end{definition}

\begin{example}
For the infinite sequence $x=(1,1,2,1,2,1,1,1,1,\ldots)$ we have $\firstpass(x) = (1,2)$.
For $x=(1,1,1,\ldots)$ we have $\firstpass(x) = (1)$.
Similarly, for the sequence $\firstpass(2,9,5,3,8,2,6,2,7,9,1,6,7,1,6) = (2,9,5,3,8,6,7,1)$.
\end{example}

A declarative process gives rise to a finite set of first-passage traces that we will herein simply call traces.

\begin{definition}
Let $D=(\Sigma,\Const)$ be a declarative process.
We denote by $\Valid(D)$ the set of first-passage traces of the process $D$:
$$\Valid(D) ~=~ \{ \firstpass(\sigma) ~:~  \sigma \in \Traces(D) \}.$$
We will use the notation $\Valid_k(D)$ to represent those length-$k$ traces in $\Valid(D)$.
We also define $\valid(D):=|\Valid(D)|$.
\end{definition}

\begin{example}
Suppose $D=(\{1,2\},\{\dresp(2,1)\})$. Then we have $\Valid(D) = \{\epsilon,(1),(2,1)\}$.
\end{example}

If there are no declarative constraints, then the activities in the process are not restricted in any way and are free to happen in any order.
There is of course no requirement that an activity has to happen.
We will use the notation $\Perms_k$ for the set of permutations of the set $\{1,\ldots,k\}$.

\begin{example}\label{freeexample}
Suppose $\Sigma=\{1,\ldots,n\}$ and consider the declarative process $D=(\Sigma,\emptyset)$.
The set of valid traces of $D$ is the set of permutations of all subsets of $\Sigma$:
\begin{align*}
\Valid(D) = \{ & \left(x_{\pi(1)},\ldots,x_{\pi(k)}\right) ~:~ \\ & X=\{x_1,\ldots,x_k\} \subseteq \Sigma \mbox{ and } \pi \in \Perms_k\}.
\end{align*}
The number of these traces is
$$\valid(D) = \sum_{k=0}^n {n \choose k} k! = n! \sum_{k=0}^n \dfrac{1}{k!} \approx n! e,$$
when $n$ is large and $e \approx 2.718$.
For the case $n=3$, we have $\valid((\{1,2,3\},\emptyset))=16$ and
\begin{align*}
&\Valid((\{1,2,3\},\emptyset)) = \{ \epsilon, (1), (2), (3), (1,2), (2,1), 
\\ & ~~
(1,3),
(3,1), (2,3), (3,2), (1,2,3), (1,3,2), 
\\ &~~ 
(2,1,3), (2,3,1),
(3,1,2), (3,2,1)\}.
\end{align*}
Note that the number of traces of length $k$ for general $n$ is ${n \choose k} k!=n(n-1)\cdots (n-k+1)$.
\end{example}

\begin{example}\label{simple}
Consider the declarative process $D=(\Sigma,\Const)$ where
\begin{align*}
\Const = \{ ~ & \dsucc(1,2), \dprec(1,3), \dresp(3,4), \\ & \drespexist(2,5), \dnotsucc(4,5)\}.
\end{align*}
The set of valid traces is
\begin{align*}
& \Valid(D) = \{ 
\epsilon,
(5) ,
(1, 2, 5) ,
(1, 5, 2) ,
(5, 1, 2) , 
\\ & ~~
(1, 2, 3, 5, 4),
(1, 2, 5, 3, 4) ,
(1, 3, 2, 5, 4), 
(1, 5, 2, 3, 4) ,
\\ & ~~
(1, 3, 5, 2, 4) ,
(1, 5, 3, 2, 4) ,
(1, 3, 5, 4, 2) ,
(1, 5, 3, 4, 2), 
\\ & ~~
(5, 1, 2, 3, 4) ,
(5, 1, 3, 2, 4) ,
(5, 1, 3, 4, 2)
\} .
\end{align*}
\end{example}

\begin{example}\label{greekgods}
As an illustrative example let us start with a plain text description of the policy making style of the 12 gods of the ancient Greek Olympian pantheon.
Much of this plain text description of the Olympian's decision-making approach would be easily recognisable by modern day public administration and public policy practitioners.

\begin{quote}
``The council of the Olympian gods and goddesses made collective decisions with input from an expert panel, which consisted of Zeus (the president of the gods), Athena (the goddess of wisdom), Hermes (the god of information and commerce), and any other god whose area of expertise would be pertinent to the subject in question. These meetings were problem-oriented participatory sessions, characterized by intense discussions and searches for best solution. The gods' decisions were persuasively communicated to mortals and powerfully implemented with follow-up reports.'' (Zanakis et al.~\cite{ZANAKIS})
\end{quote}
This Olympian policy making process can be re-formulated as the declarative process graph of activities and constraints in Fig.~\ref{ggv1}.
Briefly, the numbered activities encoding the possible decision paths in Fig.~\ref{ggv1} can be summarized as follows.
\begin{enumerate}
\item[(1)]  Identify the problem or thematic policy domain requiring attention and (2) convene the Olympian pantheon of 12 gods.
\item[(3)]  Consult the databank maintained by Hermes, the god of informatics, collecting all relevant information (5), and search for solutions through an intense dialogue of the gods (6) whilst consulting all stakeholder gods in the policy decision (4).
\item[(7)]  Propose alternative solution options and select the best solution and plan of action.
\item[(8)]  Announce the decision of the gods through the Oracle and send Peitho, the goddess of persuasion to get (9) buy-in to the decision from the mortals.
\item[(10)] Implement the decision of the gods via thunderbolts and lightening under the supervision of Hermes who follows up with a report of the outcome and (11) updates his databank.
\end{enumerate}

\begin{figure}[!h]
\label{ggv1}
\begin{tikzpicture}[outline/.style={draw=#1,thick,fill=#1!10, minimum height=0.75cm,minimum width=1.5cm},
                    outline/.default=black,
					haup/.style={thick,->}]
\newcommand{\tbox}[1]{\begin{minipage}[t]{1.5cm} \raggedright \tiny\bf #1\end{minipage}}
\def\myscl{3}
\newcommand{\myarr}[3]{\draw [haup] (#1) to node[fill=white] {#3} (#2);}
\newcommand{\myoarr}[3]{\draw [haup] (#1) to node[fill=white] {#3}  (#2);}
\node [outline] (N1) at (0,4) {\tbox{1. problem identified}};
\node [outline] (N2) at (1*\myscl,4) {\tbox{2. convene Olympian pantheon}};
\node [outline] (N3) at (0,6) {\tbox{3. consult Hermes databank}};
\node [outline] (N4) at (1*\myscl,6) {\tbox{4. consult stakeholder gods}};
\node [outline] (N5) at (2*\myscl,6) {\tbox{5. collect relevant information}};
\node [outline] (N6) at (2*\myscl,4) {\tbox{6. intense dialogue of the gods}};
\node [outline] (N7) at (1*\myscl,2) {\tbox{7. identify best solution}};
\node [outline] (N8) at (2*\myscl,2) {\tbox{8. Oracle announces decision}};
\node [outline] (N9) at (2*\myscl,0) {\tbox{9. get mortal buyin}};
\node [outline] (N10) at (1*\myscl,0) {\tbox{10. issue thunder- bolts and lightening}};
\node [outline] (N11) at (0,0) {\tbox{11. update Hermes databank}};
\myoarr{N2}{N4}{\tiny$\dprec$}
\myoarr{N2}{N3}{\tiny$\dprec$}
\myoarr{N2}{N5}{\tiny$\dprec$}
\myoarr{N2}{N7}{\tiny$\dsucc$}
\myoarr{N8}{N10}{\tiny$\dsucc$}
\myoarr{N8}{N9}{\tiny$\dprec$}
\myarr{N1}{N2}{\tiny$\dsucc$}
\myarr{N2}{N6}{\tiny$\dprec$}
\myarr{N7}{N8}{\tiny$\dsucc$}
\myarr{N10}{N9}{\tiny$\dnotsucc$}
\myarr{N10}{N11}{\tiny$\dresp$}
\end{tikzpicture}
\caption{The declarative workflow for Example~\ref{greekgods}}
\end{figure}
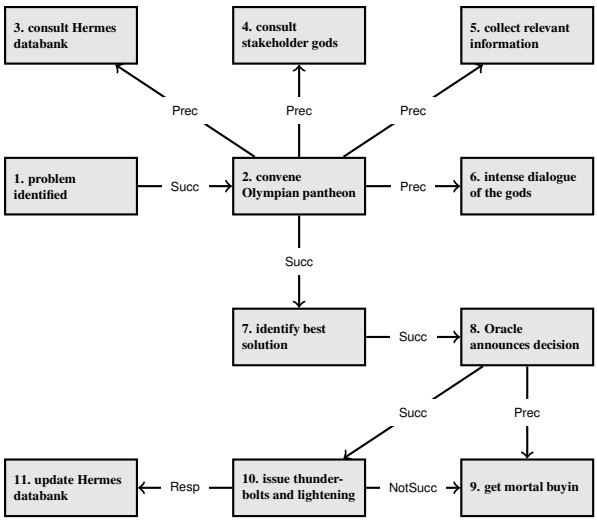

With these activities now assigned labels in the set $\{1,2,\ldots,11\}$, we may now model this process
as the declarative process
$D=(\Sigma,\Const)$ where $\Sigma=\{1,2,\ldots,11\}$ and
\begin{align*}
& \Const =
\{
\dsucc(1,2),
\dprec(2,3),
\dprec(2,4),
\dprec(2,5),
\\ & ~~~
\dprec(2,6),
\dsucc(2,7),
\dsucc(7,8),
\dprec(8,9),
\\ & ~~~
\dsucc(8,10),
\dnotsucc(10,9),
\dresp(10,11)
\}.
\end{align*}
The declarative workflow process diagram is illustrated in Figure~\ref{ggv1}.
The trace $(1, 2, 3, 7, 8, 4, 10, 11)$ is in $\Valid(D)$, and $\Valid(D)$ has size 7367.
\end{example}

\section{Measuring declarative process diversity}
Given two declarative processes $D_1=(\Sigma_1,\Const_1)$ and $D_2=(\Sigma_2,\Const_2)$, how is it possible to compare these two processes
in a way so as to measure the diversity of the processes?
This is a very general question and to approach it we must be more specific about the properties of any such measure.

Consider a general declarative process $D=(\Sigma,\Const)$.
If only one sequence of activities of $\Sigma$ may occur that satisfies $\Const$, then this is not very diverse in the sense that every activity can hold up completion of the process.
However, if any sequence of activities may occur that result in $\Const$ being satisfied, then since these can be accomplished in any order,
all activities that can happen will happen independently of one-another.
In this sense the constraints $\Const$ are satisfied at the earliest opportunity.
This leads us to the following heuristic that claims a measure of diversity of such a process should be an increasing function of the number of valid traces for that process.

\begin{heuristic}
If $D_1=(\Sigma,\Const^{(1)})$ and $D_2=(\Sigma,\Const^{(2)})$ are two declarative processes on the same set $\Sigma$,
then $D_1$ is at least as efficient as $D_2$ if $\valid(D_1) \geq \valid(D_2)$. We thus have
$$\effect(D) \propto f(\valid(D))$$
for some weakly increasing function $f$.
\end{heuristic}

In attempting to compare two declarative processes the issue of scalability arises.
If one process comprises two activities, and another comprises 100 activities,
then it makes little sense to simply compare some weakly increasing function of the number of valid traces of each of these processes.
The declarative process that gives the largest number of valid traces on an activity set $\Sigma$ is $D'=(\Sigma,\emptyset)$ given in Example~\ref{freeexample}.
It may be the case that certain constraints must always hold in any consideration, for example that some activity $a$ is in a trace, or that a trace is non-empty, and so forth.
With this in mind, we imagine that there is some subset of minimal constraints, $\MinConst \subseteq \Const$, against which we will be comparing our process $D$.
The process $D'$ corresponds to $\MinConst=\emptyset$.

Let us adopt the following piece of notation: given a declarative process $D=(\Sigma,\Const)$ and a minimal constraint set $\MinConst \subseteq \Const$, let
$D_{\MinConst} = (\Sigma,\MinConst)$.

Thus given a general declarative process $D=(\Sigma,\Const)$ with minimal constraint set $\MinConst$, the largest that $\valid(D)$ may be is $\valid(D_{\MinConst})$.
It therefore makes sense to scale the diversity by some function of the largest number of valid traces that may appear with respect to the processes that satisfies the set of minimal constraints $\MinConst$.
It is too restrictive to set $f(\valid(D)) = g(\valid(D)/\valid(D_{\MinConst}))$ as this restricts further heuristic properties for these processes to be incorporated.
While this is the simplest possible scaling and making models as simple as possible is a desirable goal, there is no reason for it to be {\it{a priori}} better that other scaling functionals.

Thus we assume the more general form for the scaling $$f(\valid(D)) = \dfrac{g(\valid(D))}{g(\valid(D_{\MinConst}))}$$ for some function $g$.
As $f$ is a weakly increasing function, $g$ too must be a weakly increasing function.
This assumption means that 1 is the maximum value $f$ can achieve over all declarative processes.

\begin{heuristic}
Suppose that $D=(\Sigma,\Const)$ is a declarative process with minimal constraint set $\MinConst$.
Then the diversity of $D$ should satisfy the relation
$$\effect(D) \propto \dfrac{g(\valid(D))}{g(\valid(D_{\MinConst}))}$$
for some weakly increasing function $g$.
\end{heuristic}

It would be difficult to use this heuristic in some practical manner without knowing further properties of $g$.
The function $g$ is not a direct measure of diversity, but represents the weight attached to the number of valid traces of a process.
Let us briefly consider processes that have 1, 10, 100, and 1000 valid traces.
A process having 1 trace is necessary for this process to realistically model some policy process, and a process having 2 traces is certainly better than a process that only has one trace.
However, we would consider a process that has 101 traces to be better, but only marginally, to a process that has 100 traces.

The simplest function that represents this situation is the function $g$ whose rate of change is inversely proportional to its argument, i.e. satisfies the differential equation $g'(x) \approx k/x$.
In order for the general solution to this, $g(x) = k\ln(x)+c$ for constants $k,c$, to represent our situation we must have $k>0$.
If there is a single valid trace for some process $D$, then we will have $g(1) = c$.
In comparing this to the free process $D_{\emptyset}$ which will have more valid traces than $D$, the diversity is thus
$\dfrac{c}{k \ln (\valid(D_{\emptyset}))+c}$.
In order to choose a sensible value of $c$ to represent this scenario, we set $c=0$ so that the diversity in this very restrictive case is 0, compared to $1$ in the case that $\valid(D) = \valid(D_{\emptyset})$.
Thus

\begin{heuristic}
Let $D=(\Sigma,\Const)$ be a declarative process with minimal constraint set $\MinConst$.
Then a sensible choice of the function $g$ that models the reducing benefit of more valid traces as the number of these valid traces increases is $g(x) = k\ln(x)$ for some positive constant $k$.
\end{heuristic}

These heuristics, when taken together, suggest the following as a measure of the diversity of a declarative process:

\begin{definition}\label{firstdef}
Let $D=(\Sigma,\Const)$ be a declarative process with minimal constraint set $\MinConst$.
Then a measure of the diversity of $D$ is
$$\effect(D) = \dfrac{\ln(\valid(D))}{\ln(\valid(D_{\MinConst}))}.$$
\end{definition}

It may be the case that the relative preferences for an increase in number of valid traces is proportional to some other weakly decreasing function of $x$.
However, we have not found any compelling motivation from the examples we have been considering for this to be the case.

An example of a minimal constraint set one might see in a declarative process that models some policy process is
$$\MinConst ~=~ \{ \dparticipation(a_{10}), \dinitial(a_3), \dfinal(a_{40})\}.$$
In the event that the minimal constraint set is empty, then Definition~\ref{firstdef} can be written more explicitly:

\begin{definition}\label{seconddef}
Let $D=(\Sigma,\Const)$ be a declarative process with $\MinConst=\emptyset$.
Then a measure of the diversity of $D$ is
$$\effect(D) = \dfrac{\ln(\valid(D))}{1+\ln( |\Sigma|!)}.$$
\end{definition}

\begin{figure}\label{effillustration}
\centerline{
\begin{tikzpicture}[scale=0.9]
	\small
    \begin{axis}[grid=both,xmax=1,xmin=0,ymax=1,ymin=0, samples=500,xlabel=$x$]
        \addplot[myonegrey, ultra thick] {log10(10*x)/log10(10)} node[below right,pos=0.2]{\tiny$10$};
        \addplot[mytwogrey, ultra thick] {log10(100*x)/log10(100)} node[below right,pos=0.12]{\tiny$100$};
        \addplot[mythreegrey, ultra thick] {log10(1000*x)/log10(1000)} node[below,pos=0.1]{\tiny$1000$};
        \addplot[myfourgrey, ultra thick] {log10(10000*x)/log10(10000)} node[above left,pos=0.1]{\tiny$10000$};
    \end{axis}
    \end{tikzpicture}
}
\caption{Illustration of $\effect$ for several $\valid(D_{\MinConst})$ values (the line labels)
with the quantity $x$ representing the proportion $\valid(D)/\valid(D_{\MinConst})$.
For example, if $D$ is a declarative process having $|\Sigma|=7$, 8, and 9 activities, respectively,
then $\valid(D_{\emptyset})$ is $13700$, $109601$, and 986410, respectively.
}
\end{figure}
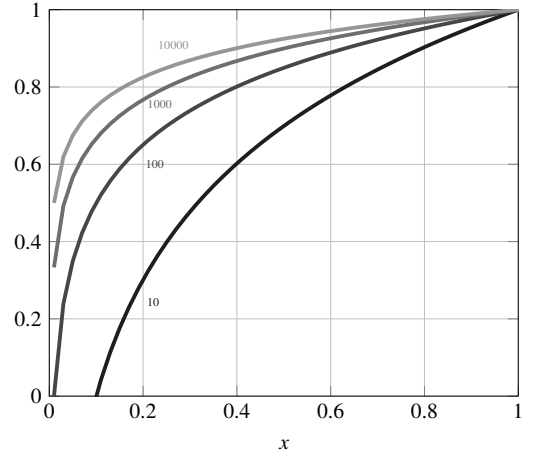

Figure~\ref{effillustration} illustrates the measure $\effect$ for several different values of $\valid(D_{\MinConst})$ (these are the values beside the coloured lines),
and the value of $x$ is the proportion $\valid(D)/\valid(D_{\MinConst})$.

\begin{example} In each of the following we assume the minimal constraint set is empty.
\begin{enumerate}
\item[(a)] For the free declarative process $D$ in Example~\ref{freeexample}, we have $\effect(D) = 1$.
\item[(b)] For the declarative process $D$ in Example~\ref{simple}, we have $\effect(D) = 0.479065690$.
\item[(c)] For the declarative process $D$ in Example~\ref{greekgods}, we have $\effect(D) =  0.481278656$.
\end{enumerate}
\end{example}

Let $D$ be a declarative process. If $\Valid(D)$ is non-empty, i.e. there is at least one valid trace (which could be the empty trace), then metric $\effect(D)$ takes values in the closed interval $[0,1]$.

In the event that $\Valid(D)$ is empty then $\effect(D)$ is not defined. There is no issue with this as it is assumed that
$D$ is a process that models a policy that is realizable. If a policy process exists that has no valid traces, then
this is a sign that the constraints defining the process are inconsistent with one another.

If there is only one activity in the set $\Sigma$, then the denominator of $\effect(D)$ will be zero and it will not be defined. Again, this is not an interesting case or one to cause alarm as the model of a policy process that consists of one activity is essentially trivial. 
(Any constraints of such a model would be unitary constraints such as `activity 1 happens' or `activity 1 does not happen'.)

The primary use of this metric is for comparing combinatorial diversity of two (of many) declarative processes that can be on completely different activity sets.
Let us explicitly mention that  there is no reason for the number of activities in each of the processes to be the same.
We can conclude the process $D_1$ is more combinatorial diverse that process $D_2$ if $\effect(D_1)> \effect(D_2)$.

Can we attribute a meaning to a particular value of $\effect(D)$?
For example, if $\effect(D)  \approx k$ then what can we infer about the process?
From the definition of $\effect$, this means that is satisfies the power-law relation:
$\valid(D))= \valid(D_{\MinConst})^{k}$.
So, for example, if $\effect(D)=0.5$ then this corresponds to those declarative processes for which the
number of valid traces is the square root of the number of traces of the associated minimal (or free) process.

\section{Measuring a specified `goodness' in valid traces}
Given a declarative process $D=(\Sigma,\Const)$, the set $\Valid(D)$ is the set of those valid traces illustrating how the activities of the
process happen in relation to one another.
It may be the case that some activities are deemed desirable or good.
In order to attribute a meaning to these notions that can be used in some quantification, we must be able to specify whether each trace
in $\Valid(D)$ is 'good' or 'not-good', and we will do this by specifying a collection of declarative constrains $\GConst$.
A trace will be called {\it{good}} if it satisfies all those constraints in $\GConst$.

\begin{definition}
Let $D=(\Sigma,\Const)$ be a declarative process. Let $\GConst$ be a collection of declarative constraints.
A trace $\sigma \in \Valid(D)$ will be called {\it{good}} if $\sigma \models \GConst$.
Let $\GValid(D) = \{ \sigma \in \Valid(D) ~:~ \sigma \models \GConst\}$ and $gvalid:=|\GValid|$.
\end{definition}

\begin{example}\label{simplegood}
Consider the declarative process of Example~\ref{simple}.
Let us suppose that our notion of goodness is that activity 4 occurs and that activity 3 occurs before 2 (should they happen at all).
We thus have $\GConst = \{ \dparticipation(4), \dsucc(3,2)\}$.
In this case we have
\begin{align*}
& \GValid(D) = \{ 
(1, 3, 2, 5, 4),
(1, 3, 5, 2, 4) , 
\\ & ~~
(1, 5, 3, 2, 4) ,
(1, 3, 5, 4, 2) ,
(1, 5, 3, 4, 2), 
\\ & ~~
(5, 1, 3, 2, 4) ,
(5, 1, 3, 4, 2)
\} .
\end{align*}
\end{example}

\begin{definition}
Let $D=(\Sigma,\Const)$ be a declarative process. Let $\GConst$ be a collection of declarative constraints.
Let us define two goodness metrics of the process $D$ with respect to $\GConst$:
\begin{align*}
\goodness (D,\GConst) &= \dfrac{\gvalid}{\valid}, \mbox{ and }\\ 
\loggoodness(D,\GConst) &= \dfrac{ \ln \gvalid}{\ln \valid}.
\end{align*}
\end{definition}

\begin{example}
Applying the previous definition to Example~\ref{simplegood} we have
$ \goodness (D,\GConst)   = {7}/{16} = 0.4375$ and 
$\loggoodness(D,\GConst)  = \ln (7)/\ln( 16) = 0.70183.$
\end{example}

Just as with the $\effect$ metric, both of these metrics can be used to compare a collection of different declarative processes each with their own respective goodness constraints.

In using these two goodness metrics, we envisage a declarative process $D$ that models some policy process
and some constraint $\GConst$ against which every trace $\sigma \in \Valid(D)$ can be classified as `good' or `not good'.
The metric $\goodness$ takes values in the closed interval $[0,1]$ and gives the proportion of valid traces that are good amongst all valid traces.
The metric $\loggoodness$ also takes values in the closed interval $[0,1]$ and
produces a number $k$ that relates the two quantities in terms of a power law: number of good traces $\sim$ (number of traces)$^{k}$.

The question of which metric to choose is of course a subjective one.
If we are simply interested in the proportion of good traces to valid traces then the metric $\goodness$ is, by definition, the best choice.
However, if the doubling of the number of good traces should represent something strictly less than a two-fold increase in the levels of goodness, then the $\loggoodness$ metric is the better choice.

The second metric $\loggoodness$ is not defined for two different degenerate cases: when there are no good traces in the list of valid traces (this would imply the numerator contains the undefined term $\ln(0)$), and when there is only one valid trace (this could cause a denominator of 0).

\section{The entropy of random traces}
\mps{Expand section title}

Consider the declarative process $D=(\Sigma,\Const)$ with minimal constraint set $\MinConst$.
Let us consider the set of valid traces for this process, $\Valid(D)$.
Recall that as we are dealing with first-passage traces, the set $\Valid(D)$ contains no duplicate sequences, and we have
$\Valid(D) \subseteq \Valid(D_{\emptyset})$.

The outcome of a declarative process $D$ is a trace. 
Let $X_1:=X_1(D)$ be the random variable that represents the outcome of the process $D$.
In the absence of further information, all valid traces are equally likely and we have
$$\P(X_1= \sigma) ~=~
\begin{cases}
\dfrac{1}{\valid(D)} & \mbox{ if }\sigma \in \Valid(D)\\[1em]
0 & \mbox{ if }\sigma \not\in \Valid(D).
\end{cases}
$$

Let us observe that the entropy of this random variable $X_1$ is simply calculated as
\begin{align*}
H(X_1) = - \sum_{\sigma} \P(X_1= \sigma) \ln (\P(X_1= \sigma)) = \ln(\valid(D)).
\end{align*}
This quantity, known both as the `max-entropy' and as the Hartley function, is the largest value that any probability measure on a set of size $\valid(D)$ may achieve.
This fact grows in importance once we realise that it is the same quantity that appears in the numerator of $\effect$ in Definition~\ref{firstdef}.
Indeed, we can re-write the measure in terms of the entropy as
$$\effect(D) = \dfrac{H(X_1(D))}{H(X_1(D_{\MinConst}))}.$$

\section{A metric motived by pattern distribution in logs}
The set of valid traces of a declarative process will allow for many permutations of particular actions at particular positions.
It will also be the case that there are certain subsequences (or patterns) of events that simply cannot happen due to the constraints.
We require a metric that reflects the level of `freeness' with respect to patterns that may or may not happen, and this metric
must take into account the permutive aspect of our considerations.

More formally, consider a general declarative process $D=(\Sigma,\Const)$ with $L=\Valid(D)$.
Let us fix a pattern length $n$ that we will think of as the `resolution' of our pattern analysis.
We wish to derive a measure of the pattern complexity of $L$, and will refer to it as $\pc_n=\pc_n(L,\Sigma)$.
The traces in $L$ are sequences of unique entries from the set $\Sigma$, 
The metric $\pc_n$ should be independent of the labels of $\Sigma$.

\begin{heuristic}\label{permok}
If $\pi(\Sigma)$ is a permutation of the set $\Sigma$, then
we require $$\pc_n(L,\Sigma) = \pc_n(L_{\pi},\Sigma)$$
where $L_{\pi}$ is the log $L$ with every entry $x_i$ in each trace $x$ replaced with $\pi(x_i)$.
\end{heuristic}

In order to introduce the notion of a (permutation) pattern, we must assume some total order ($\preceq$) on $\Sigma$.
Let $x=(x_1,\ldots,x_t)$ be a sequence where $x_i\in \Sigma$ and there are no duplicate entries, i.e. all entries of $x$ are unique.
A subsequence $x'=(x_{i_1},\ldots,x_{i_k})$ of $x$ is an occurrence of the pattern $p=(p_1,\ldots,p_k) \in \Perms_k$
if they are order isomorphic: i.e. the smallest (with respect to the order $\preceq$) entry of $x'$ is in the same position as the smallest (with respect to the order $\leq$) entry of $p$, the second smallest
entry of $x'$ is in the same position as the second smallest entry of $p$, and so on.

Given any subsequence $x'$ of $x$ having length $k$, it will the order isomorphic to precisely one permutation $p \in \Perms_k$.
In such a case we say that {\it{$x'$ is an occurrence of the pattern $p$ in $x$}} and or that {\it{$x$ contains the pattern $p$.}}
Given a pattern $p \in \Perms_k$, let $p(x)$ be the number of occurrences of the pattern $p$ in $x$.

\begin{example}
Let $x=(5,9,2,6,20,3,12,18)$. Then $x'=(9,6,20,3,18)$ is an occurrence of the pattern $p=(3,2,5,1,4)$ in $x$.
Similarly, $x''=(2,18)$ is an occurrence of the pattern $q=(1,2)$ in $x$.
Also $p(x) = 1$ and $q(x) = 19$.
\end{example}

\begin{definition}
Given a log $L$ and integer $n$, let $Y$ be the pattern that results from choosing a random trace of $L$ and selecting a random length-$n$ subsequence of that trace.
\end{definition}

Let $(P(i))_{i=1}^{n!}$ be a listing of the elements of $\Perms_n$ in lexicographic order.
We have
$$p_{\alpha} ~:=~ \Prob(Y=\alpha) ~=~ N^{(\pi)}(L)/N^{(n)}(L)$$
where $N^{(\pi)}(L)$ be the number of occurrences of a pattern $\pi\in\Perms_n$ in the set $L$ and let $N^{(n)}(L)$ be their sum:
\begin{align}
\label{mynd}
N^{(\pi)}(L) = \sum_{x \in L} \pi(x)\mbox{ and }
N^{(n)}(L)  = \sum_{\pi \in \Perms_n} N^{(\pi)}(L).
\end{align}

Our measure of pattern diversity, $\pc_n(L)$, will depend on these probabilities $p_{\alpha}$.
It must also be such that any permutation of the values will not change the metric due to Heuristic~\ref{permok}.
Thus we have
\begin{heuristic}
For any permutation $\pi \in \Perms_n$, the $n$-pattern diversity should be invariant of the action of $\pi$ on
the distribution of pattern occurrences:
\begin{align*}
\pc_n(L) &= f(p_{P(1)},\ldots,p_{P(n!)})\\ &= f(p_{\pi(P(1))},\ldots,p_{\pi(P(n!))}).
\end{align*}
\end{heuristic}

Reasoning further about how this pattern diversity metric should behave, the extreme values are straightforward to characterize:

\begin{heuristic}
The function $f$ should attain a maximum when $p_{P(1)}=p_{P(2)}=\ldots=p_{P(n!)}$ since this would indicate that the $n$-patterns in the log traces are as evenly distributed (and therefore permutationally diverse) as they can be.
If all $n$-patterns in $L$ are the same pattern, then this means the $n$-patterns in the log traces are as undiverse as is possible, and the function $f$ should take the value 0 in this case. Note that this will mean exactly on of the $p_{\alpha} = 1$ and all others are 0.
\end{heuristic}

These heuristics provide a compelling argument for choosing the entropy of the random variable $Y$ to be the function $f$ (on which $\pc_n$ is based).
They form a subset of the axioms proposed by Shannon in \cite{SHANNON} and for which he showed the Shannon entropy was the unique solution.

\begin{definition}
Let $L$ be a set of sequences where every sequence contains only distinct entries, and let $n$ be an integer representing pattern length.
Let $N^{(\pi)}(L)$ be the number of occurrences of a pattern $\pi\in\Perms_n$ in the set $L$ and let $N^{(n)}(L)$ be their sum (see Eqn.~\ref{mynd}).
Set $p_{\pi}:=p_{\pi}(L) = N^{(\pi)}(L)/N^{(n)}(L)$ and define the {\it{$n$-pattern diversity of $L$}} to be
$$\pc_n(L,\Sigma) ~:=~ -\sum_{\pi\in\Perms_n} p_{\pi}  \ln  (p_{\pi}).$$
The $n$-permutation entropy has 0 and $\ln (n!)$ as its minimum and maximum value, respectively.
To scale these entropies we introduce the
{\it{normalized $n$-permutation entropy}}
$$\npc_n(L,\Sigma) ~:=~ \dfrac{\pc_n(L,\Sigma)}{\ln (n!)}.$$
\end{definition}

\begin{example}
\begin{enumerate}
\item[]
\item[(i)]  For the free declarative process $D$ of Example~\ref{freeexample} on $n$ activities, all permutations of all subsets of the activity set are valid traces.
Thus all of the probabilities $p_{\alpha} = 1/n!$ and $\pc_n(\Valid(D),\Sigma) = \ln (n!)$ and $\npc_n(\Valid(D),\Sigma)=1$.
\item[(ii)] For the declarative process $D$ in Example~\ref{simple}:\\[0.5em]
\centerline{
\small
\begin{tabular}{|l|c|c|} \hline
$n$ & $\pc_n$ & $\npc_n$ \\ \hline
3 & 1.506127592 & 0.840585814\\
4 & 2.335683250 & 0.734941374\\
5 & 2.397895273 & 0.500866717\\ \hline
\end{tabular}
}
\item[(iii)] For the declarative process $D$ in Example~\ref{greekgods}:\\[0.5em]
\centerline{
\small
\begin{tabular}{|l|c|c|} \hline
$n$ & $\pc_n$ & $\npc_n$ \\ \hline
3 & 1.360117844 & 0.759096222 \\
4 & 2.304788489 & 0.725220091 \\
5 & 3.277594190 & 0.684616155 \\ \hline
\end{tabular}
}
\end{enumerate}
\end{example}

The normalised metric allows us to compare the diversity seen between completely different processes and is invariant under a relabelling of the activities. The higher the value the more diverse they are in terms of the $n$-patterns.

The metrics are well-defined for values of $n$ between 1 and the length of the longest trace in $\Valid(D)$.
It would be extremely unusual to find a declarative process that models a policy that has a pattern diversity of $0$ or $1$.
Such instances should be scrutinized to ensure that the list of valid traces is not something trivial (such as a single trace).
We have strong reasons to suspect that length 3, 4 and 5 patterns will produce the most interesting metrics for comparative purposes.

A related concept, in spirit, is `permutation entropy'~\cite{brandtpompe}.
Permutation entropy is an analytical tool for studying patterns in time series data in statistics that utilizes the more restrictive notion of `consecutive pattern'.
Interestingly, it has been applied to a wide variety of time series data to detect temporal changes with a view to predicting stock market behavior, detecting obstructive sleep apnea, and predicting epilepsy.

\section{Conclusion}
We have used heuristic reasoning to derive metrics that can be used to compare policy processes through combinatorial considerations.
This provides a theoretically justifiable method that does not rely on {\it{a priori}} quantitative information.


\begin{thebibliography}{00}
\bibitem{brandtpompe} C. Brandt \& B. Pompe. Permutation entropy -- a natural complexity measure for time series. {\it Phys. Rev. Lett.} 88, 174102, 2002.
\bibitem{DAVENPORT} T.H. Davenport. {\it{Process Innovation: Reeingineering Work through Information Technology}}. Boston, MA: Harvard Business School Press, 1993.
\bibitem{ltlfinite} V. Fionda and G. Greco.  LTL on Finite and Process Traces: Complexity Results and a Practical Reasoner. {\it{J. Artificial Intelligence Res.}} {\bf{63}}:557--623, 2018.
\bibitem{HEITSCH} S. Heitsch, D. Hinck \& M. Martens. A New Look into Garbage Cans -- Petri Nets and Organisational Choice. {\it{Proceedings of the AISB'00 Symposium on Starting from Society -- the Application of Social Analogies to Computational Systems}} 51--60, 2000.
\bibitem{LASSWELL} H.D. Lasswell. {\it{The Decision Process: Seven Categories of Functional Analysis}}. College Park, MD, University of Maryland Press, 1956.
\bibitem{MELCHER} J. Melcher. {\it{Process Measurement in Business Process Management: Theoretical Framework and Analysis of Several Aspects.}} KIT Scientific Publishing, 2012.
\bibitem{MENDLING} J. Mendling. {\it{Metrics for Process Models: Empirical Foundations of Verification, Error Prediction, and Guidelines for Correctness}}. Heidelberg, Germany: Springer, 2008.
\bibitem{finiteltlone} M. Pesic, D.  Bosnacki \& W. van der Aalst. Enacting Declarative Languages Using LTL: Avoiding Errors and Improving Performance. {\it{Proc. of SPIN}}, 146--161, 2010.
\bibitem{render} S. Redner. {\it{A Guide to First-Passage Processes}}. Cambridge: Cambridge University Press, 2001. 
\bibitem{GONZALEZ} L. S\'anchez Gonz\'alez, F. Garc\'ia Rubio, F. Ruiz Gonz\'alez \& M. Piattini Velthuis. Measurement in business processes: a systematic review. {\it{Bus. Process Manag. J.}} {\bf{16}}:114--134, 2010.
\bibitem{SHANNON} C.E. Shannon. A Mathematical Theory of Communication. {\it{The Bell System Technical Journal}} {\bf{27}}(3):379--423, 1948.
\bibitem{vda} W. van der Aalst. {\it{Process Mining: Data Science in Action}}. Second edition. Springer, 2016.
\bibitem{vda2014} W. van der Aalst. {\it{Process mining: discovery, conformance and enhancement of business processes}}. Berlin, Heidelberg: Springer, 2014.
\bibitem{declarelanguage} W. van der Aalst, M. Pesic \& H. Schonenberg. Declarative Workflows: Balancing Between Flexibility and Support. {\it{Computer Science -- Research and Development}} {\bf{23}}(2), 99--113, 2009.
\bibitem{WEIBLE} C.M. Weible \& P.A. Sabatier. {\it{Theories of the Policy Process}}. Fourth edition. Boulder, CO: Westview Press, 2017.
\bibitem{ZANAKIS} S.H. Zanakis, S. Theofanides, A.N.  Kontaratos \& T.P. Tassios. Ancient Greeks' Practices and Contributions in Public and Entrepreneurship Decision Making. {\it{Interfaces}} {\bf{33}}:72--88, 2003.
\end{thebibliography}
\end{document}